\pdfoutput=1

\documentclass[11pt]{article}

\newcommand\blfootnote[1]{%
  \begingroup
  \renewcommand\thefootnote{}\footnote{#1}%
  \addtocounter{footnote}{-1}%
  \endgroup
}

\usepackage[preprint]{acl}

\usepackage{times}
\usepackage{latexsym}
\usepackage{blindtext}

\usepackage[T1]{fontenc}

\usepackage[utf8]{inputenc}

\usepackage{microtype}

\usepackage{inconsolata}

\usepackage{graphicx}

\title{Can Agent Conquer Web? Exploring the Frontiers of ChatGPT Atlas Agent in Web Games}

\author{Jingran Zhang$^{1*}$  \quad Ning Li$^{1*}$ \quad Justin Cui$^{2}$ \\
$^{1}$ UC San Deigo, $^{2}$ UCLA
}

\begin{document}
\maketitle

\blfootnote{* indicates equal contributions.}

\begin{figure*}[ht]
    \centering
    \includegraphics[width=\textwidth]{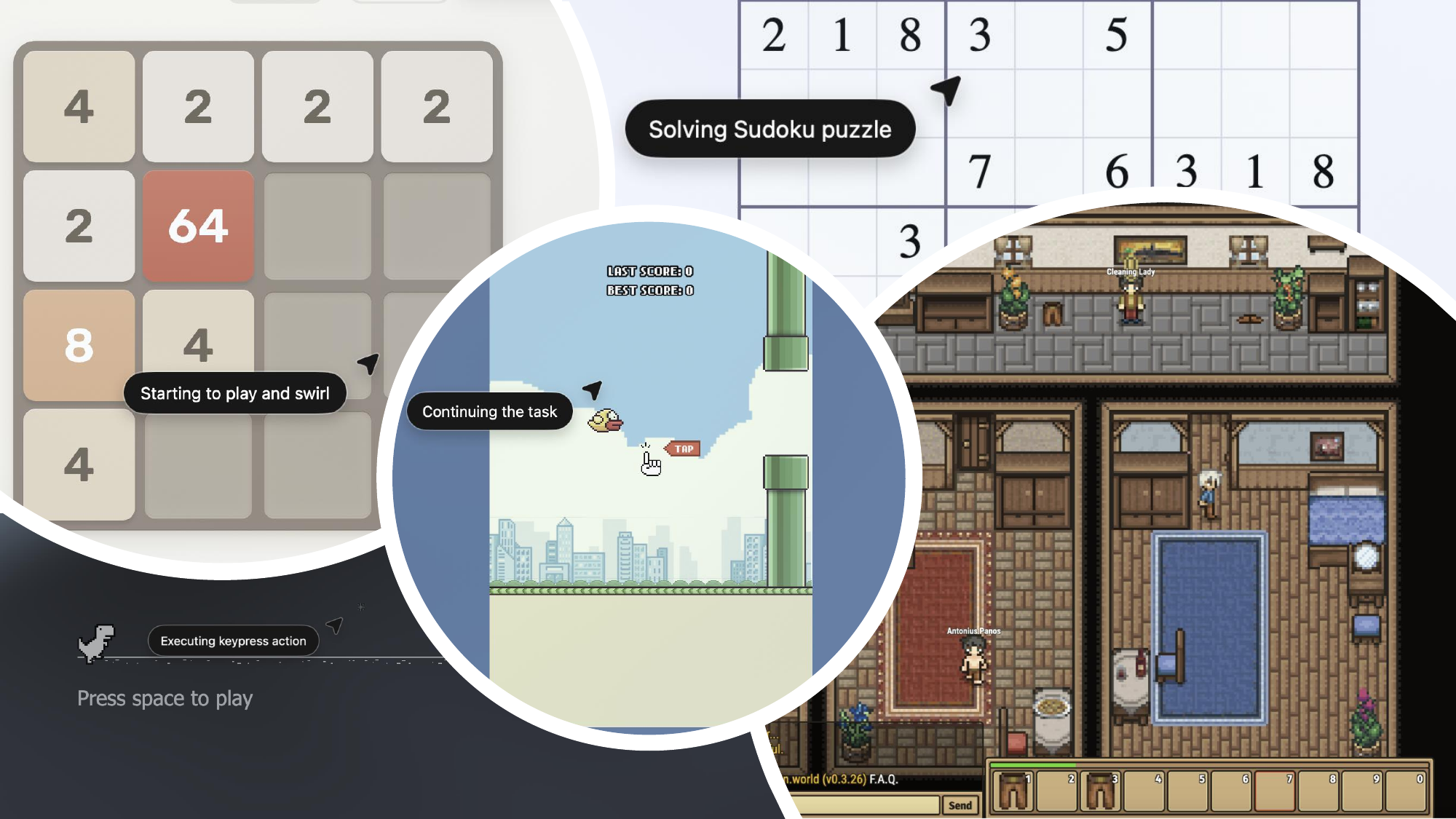}
    \caption{ChatGPT Atlas being evaluated across different web games. The agent performs 
     strategic moves in 2048 (Top-Left), logical placements in Sudoku (Top-Right),
     key presses for timing in T-Rex Runner (Bottom-Left), rapid taps for continuous control in Flappy Bird (Center), and exploration, navigation, game-understanding in Stein.world (Bottom-Right).
    }
    \label{fig:pipeline}
\end{figure*}

\begin{abstract}

\noindent OpenAI's ChatGPT Atlas introduces new capabilities for web interaction, enabling the model to analyze webpages, process user intents, and execute cursor and keyboard inputs directly within the browser. While its capacity for information retrieval tasks has been demonstrated, its performance in dynamic, interactive environments remains less explored. In this study, we conduct an early evaluation of Atlas's web interaction capabilities using browser-based games as test scenarios, including Google's T-Rex Runner, Sudoku, Flappy Bird, and Stein.world. We employ in-game performance scores as quantitative metrics to assess performance across different task types. Our results show that Atlas performs strongly in logical reasoning tasks like Sudoku, completing puzzles significantly faster than human baselines, but struggles substantially in real-time games requiring precise timing and motor control, often failing to progress beyond initial obstacles. These findings suggest that while Atlas demonstrates capable analytical processing, there remain notable limitations in dynamic web environments requiring real-time interaction. The website of our project can be found at \url{https://atlas-game-eval.github.io}.
\end{abstract}

\section{Introduction}

The development of AI systems for web interaction has evolved from basic text-based assistants to more sophisticated tools that can understand and manipulate browser interfaces. OpenAI's ChatGPT Atlas represents a recent advancement in this space, offering capabilities to analyze webpages, process user instructions, and perform cursor and keyboard inputs directly within the browser environment \cite{openai2025atlas}. While the system has shown promise for straightforward tasks like information retrieval, its performance in more dynamic, interactive web environments remains less understood. This gap is notable, especially as benchmarks for generalist web agents (e.g., Mind2Web \cite{deng2023mind2webgeneralistagentweb}) and visual foundation agents (e.g., VisualAgentBench \cite{liu2024visualagentbenchlargemultimodalmodels}) have established the complexity of real-world digital tasks. While VisualAgentBench systematically compares multiple models across structured tasks, we conduct an in-depth behavioral analysis of a single deployed system, examining how it handles the unpredictability of interactive web games and the integration challenges between analytical reasoning, motor control, and contextual understanding.

Web-based games provide useful test scenarios for examining these interactive capabilities, as they offer clear performance metrics and cover diverse interaction types—from logical puzzles to real-time challenges. This approach follows a long tradition of using games as AI benchmarks, famously exemplified by AlphaGo \cite{44806}. Previous evaluations of web interaction systems have often focused on static tasks or simple navigation in constrained environments \cite{zhou2024webarenarealisticwebenvironment}, leaving open questions about how systems handle the timing and adaptation demands of highly interactive applications.

In this work, we present an early evaluation of ChatGPT Atlas using several browser games to observe its web interaction capabilities. We examine how the system handles different types of tasks, focusing on four key aspects:

\begin{enumerate}
\item \textbf{Analytical Processing}: How well does Atlas understand game rules and objectives from webpage content?
\item \textbf{Input Execution}: How accurately does it translate intentions into cursor and keyboard actions?
\item \textbf{Adaptive Behavior}: Does it adjust its approach when encountering difficulties?
\item \textbf{Contextual Understanding}: How effectively does Atlas comprehend narrative instructions and pursue multi-step objectives in text-rich environments?
\end{enumerate}

We tested Atlas on multiple games including Google's T-Rex Runner, Sudoku, 2048, Flappy Bird, and the narrative-driven RPG (Role-Playing Game) Stein.world, using both quantitative performance metrics and qualitative behavioral analysis. Our observations show that while Atlas demonstrates \textbf{capable analytical reasoning in puzzle games} like Sudoku, it encounters \textbf{significant challenges in real-time games} requiring precise timing and coordination. In RPG environments, Atlas shows heavy dependence on explicit instructions and struggles with contextual narrative understanding and autonomous objective pursuit. Interestingly, we also noted several adaptive behaviors, such as \textbf{attempting to activate cheat modes} when struggling and adjusting interaction strategies over multiple attempts.

These preliminary observations provide initial insights into Atlas's current capabilities and limitations for interactive web tasks, suggesting directions for further investigation of web-based AI systems.


\section{Related Work}

\textbf{Web Interaction Agents}
The development of AI systems for web interaction has progressed from script-based automation and traditional testing frameworks \cite{kertusha2025surveywebtestingrise} to sophisticated neural approaches. Contemporary systems increasingly leverage large language models and multimodal architectures for improved generalization across diverse web environments \cite{ning2025surveywebagentsnextgenerationai}. Recent benchmarks have established comprehensive evaluation protocols, with WebLINX focusing on multi-turn conversational navigation \cite{lù2024weblinxrealworldwebsitenavigation} and WebVoyager demonstrating end-to-end task completion using large multimodal models \cite{he2024webvoyagerbuildingendtoendweb}. However, recent critical analysis suggests potential over-optimism in reported capabilities, with significant performance gaps emerging under rigorous online evaluation \cite{xue2025illusionprogressassessingcurrent}.

State-of-the-art systems exhibit diverse architectural approaches: CogAgent specializes in GUI understanding through high-resolution visual processing \cite{hong2024cogagentvisuallanguagemodel}, while WebSight adopts a vision-first paradigm that operates purely through screenshots without HTML dependency \cite{bhathal2025websightvisionfirstarchitecturerobust}. The field is increasingly moving toward more dynamic evaluation frameworks, with benchmarks like WebCanvas addressing online environments \cite{pan2024webcanvasbenchmarkingwebagents} and RealWebAssist focusing on long-horizon assistance with real-world users \cite{ye2025realwebassistbenchmarklonghorizonweb}. WebWalker further extends evaluation to systematic web traversal tasks \cite{wu2025webwalkerbenchmarkingllmsweb}. ChatGPT Atlas represents the latest evolution in this trajectory, integrating direct browser control within a general-purpose multimodal framework to enable seamless task execution across diverse web environments.

\textbf{AI and Game Evaluation}
Games have long served as standardized probes for interactive competence, with the Arcade Learning Environment (ALE) establishing a multi-game protocol for evaluating general agents across diverse Atari tasks \cite{Bellemare_2013}. More recently, LLM/VLM-oriented benchmarks extend this paradigm to contemporary agent stacks: \textit{BALROG} evaluates agentic reasoning of language and vision-language models across challenging, long-horizon game environments with fine-grained metrics \cite{paglieri2025balrogbenchmarkingagenticllm}, while \textit{Orak} targets diverse real-world video games and emphasizes modular “agentic” components and standardized connection interfaces for reproducible evaluation \cite{park2025orakfoundationalbenchmarktraining}. Complementary open-ended frameworks such as \textit{MCU} situate evaluation in Minecraft with compositional task generation and human-aligned scoring, highlighting scalability and difficulty control in sandbox settings \cite{zheng2025mcuevaluationframeworkopenended}. Parallel to benchmarking for play quality, work on automatic playtesting uses LLMs to drive coverage and defect discovery (e.g., snapshot–to–matrix perception with prompting-based action for match-3 games), foregrounding QA outcomes rather than agent interaction competence per se \cite{Zhao_2025}.

\textbf{Multimodal Web Understanding}
Multimodal web understanding has increasingly emphasized fine-grained grounding and layout-aware reasoning under real websites and realistic UI tasks. Benchmarks such as VisualWebBench target screenshot-based OCR, understanding, and element grounding across diverse, text-rich pages, revealing persistent gaps—particularly under low-resolution inputs and dense typography \cite{liu2024visualwebbenchfarmultimodalllms}. WebMMU further unifies three core web tasks—website VQA, HTML/CSS/JS code editing with functional preservation, and mockup-to-code generation—on expert-annotated real webpages, and reports that current MLLMs struggle with precise grounding, multi-step reasoning, and maintaining functionality during editing and generation, including multilingual settings \cite{awal2025webmmubenchmarkmultimodalmultilingual}. Within the broader MLLM landscape, surveys catalog evaluation axes spanning perception/understanding, cognition/reasoning, domains, and modalities, situating web-focused resources like VisualWebBench and WebMMU inside a larger benchmark taxonomy that is still evolving \cite{li2024surveybenchmarksmultimodallarge}.


Our preliminary evaluation contributes to this landscape by providing initial observations of ChatGPT Atlas's performance across different types of web-based games, focusing on practical interaction capabilities rather than specialized game-playing performance.

\section{Experiment}

\subsection{Experimental Setup}

\textbf{Technical Environment}
All experiments were conducted using ChatGPT Atlas browser (version: October 21, 2025 release) on macOS Sonoma 14.6.1. Testing occurred under standard WiFi network conditions. The evaluation utilized the "Agent Mode (Preview)" feature available to Plus, Pro, and Business users, which operates under the following constraints as specified in the release notes: no system code execution, no file system access, and no memory access. 

\textbf{Interaction Protocol}
We employed a standardized zero-shot evaluation protocol across all games. For each trial, we:
\begin{enumerate}
\item Navigated to the target game URL in ChatGPT Atlas browser
\item Opened the ChatGPT sidebar interface
\item Enabled Agent Mode (Preview)
\item Provided the initial instruction: \texttt{``Try your best to play the game until you get stuck."}
\item Recorded all interactions without additional prompting or intervention
\end{enumerate}

This protocol was designed to assess Atlas's autonomous task execution capabilities without iterative guidance or human supervision during gameplay.

\textbf{Game Specifications and Conditions}
We evaluated five web-based games as represented in Fig. \ref{fig:pipeline} representing different interaction paradigms:

\begin{itemize}
\item \textbf{2048 (Strategy/Puzzle)}: \url{https://play2048.co} - Tile merging game requiring strategic planning.
\item \textbf{Google T-Rex Runner (Reflex/Arcade)}: \url{https://chrome-dino-game.github.io} - Infinite runner demanding precise timing and obstacle avoidance.
\item \textbf{Sudoku (Logic/Puzzle)}: \url{https://www.websudoku.com?level=2} - All puzzles set to medium difficulty (level 2).
\item \textbf{Flappy Bird (Real-time Control)}: \url{https://flappybird.io/index.html} - Continuous control game requiring rhythmic tapping.
\item \textbf{Stein.world}: \url{https://Stein.world/} - A free browser MMORPG with action-combat in a 2D pixel fantasy world.
\end{itemize}

Each trial began from identical initial states: fresh browser sessions with cleared caches for consistent starting conditions. Sudoku puzzles were randomly generated but consistently of medium difficulty across all trials.


\subsection{Experimental Design}

We employed a mixed-methods evaluation approach, combining quantitative performance assessment across four structured games with qualitative behavioral analysis of a MMORPG environment.

For the quantitative assessment, we conducted 10 independent trials for each of four games featuring standardized performance metrics, as shown in Table. \ref{tab:performance}:
\begin{itemize}
\item \textbf{T-Rex Runner}: Final score (distance traveled) and obstacle clearance rate
\item \textbf{Sudoku}: Completion time with 100\% accuracy requirement
\item \textbf{Flappy Bird}: Survival score (pipes cleared)
\item \textbf{2048}: Final score and strategic progression
\end{itemize}

Complementing this quantitative analysis, we conducted a qualitative case study using the RPG (Role-Playing Games) Stein.world to examine Atlas's capabilities in open-ended, narrative-driven environments. As MMORPG lacking standardized performance metrics, this component focused on observational analysis of interaction patterns, instructional comprehension, and adaptive behaviors during initial task completion.

Human baseline data was collected from established sources: Sudoku completion times from published averages (10-12 minutes for medium difficulty), T-Rex Runner, Flappy Bird, and 2048 scores from initial human trials.






\begin{table*}[t] 
\centering
\begin{tabular}{l|c|c|c|c} 
\hline
\textbf{Game} & \textbf{Metric} & \textbf{Atlas Performance} & \textbf{Human Baseline} & \textbf{Performance Gap} \\
\hline
T-Rex Runner & Average Score & 45.5 ($\sigma$=2.92) & 388.9 ($\sigma$=325.9) & 88.3\% \\
Sudoku & Completion Time & 2m28s ($\sigma$=29s) & 10--12m & $-$75\% (faster) \\
Flappy Bird & Average Score & 0 & 2.9 (initial attempts) & 100\% \\ 
2048 & Average Score & 2242.0 ($\sigma$=1189.0) & 3463.2 ($\sigma$=2219.5) & 35.3\% \\
\hline
\end{tabular}
\caption{Empirical Performance Results Across Web Games. $\sigma$ denotes the standard deviation.}
\label{tab:performance}
\end{table*}

\subsection{Experimental Observations}

Our experiments revealed a clear \textbf{dichotomy }in ChatGPT Atlas's performance based on the motor and cognitive demands of each game. We categorize the games into three distinct types based on their interaction requirements:

\begin{itemize}
\item \textbf{High Motor-Demand Games}: Require precise timing and continuous control (T-Rex Runner, Flappy Bird)
\item \textbf{Low Motor-Demand, High Strategy Games}: Emphasize analytical reasoning with minimal real-time demands (Sudoku)
\item \textbf{Exploration-Intensive Games}: Require interface discovery and strategic adaptation (2048, Stein.world)
\end{itemize}

\paragraph{T-Rex Runner: Systematic Timing Failures}
In T-Rex Runner, Atlas achieved an average score of 45.5 points ($\sigma$=2.92), representing only 11.7\% of the human baseline performance (388.9 points). The model \textbf{failed to pass the first obstacle in 9 out of 10 trials}, with failure analysis revealing consistent late jump timing. 

During experiment, Atlas demonstrated awareness of its limitations by attempting to activate the game's \textbf{"start slower" option}. The interaction sequence was: \\
1) Mouse hover over game settings area; \\
2) Click on settings icon (when visible); \\
3) Attempt to locate and activate "start slower" or difficulty reduction.

While this adaptation strategy showed problem-solving intent, the game interface did not provide accessible difficulty controls, preventing successful execution.

\paragraph{Flappy Bird: Uncoordinated Motor Control}
Atlas demonstrated complete failure in Flappy Bird, achieving \textbf{0 points across all 10 trials} compared to human baseline scores of 1-6 points in initial attempts as shown in Table. \ref{tab:flappy_performance}. The failure mode analysis revealed erratic and uncoordinated tapping patterns without rhythmic timing relative to the bird's descent trajectory.

\begin{table}[h]
\centering
\begin{tabular}{l|c|c}
\textbf{Trial} & \textbf{Atlas Score} & \textbf{Human Baseline} \\
\hline
1 & 0 & 2 \\
2 & 0 & 2 \\
3 & 0 & 6 \\
4 & 0 & 5 \\
5 & 0 & 3 \\
6 & 0 & 2 \\
7 & 0 & 2 \\
8 & 0 & 1 \\
9 & 0 & 3 \\
10 & 0 & 3 \\
\end{tabular}
\caption{Flappy Bird performance comparison between ChatGPT Atlas and a human baseline. The values represent the number of pipes successfully passed per trial, where higher scores indicate better performance. Atlas failed to pass any pipes across all ten trials.}
\label{tab:flappy_performance}
\end{table}


  In observation, Atlas \textbf{systematically increased its click frequency} from trial to trial. However, this adaptation failed to improve performance because the clicks lacked temporal coordination with the game's physics. The model increased quantity of inputs without improving quality of timing.

\paragraph{Sudoku: Efficient Logical Reasoning}
In contrast to the motor-demand games, Atlas demonstrated exceptional performance in Sudoku, completing medium-difficulty puzzles with 100\% accuracy in an average time of 2 minutes 28 seconds ($\sigma$=29 seconds). This represents a \textbf{4.5x speed advantage over the human baseline} of 10-12 minutes.

\begin{figure*}[ht]
    \centering
    \includegraphics[width=\textwidth]{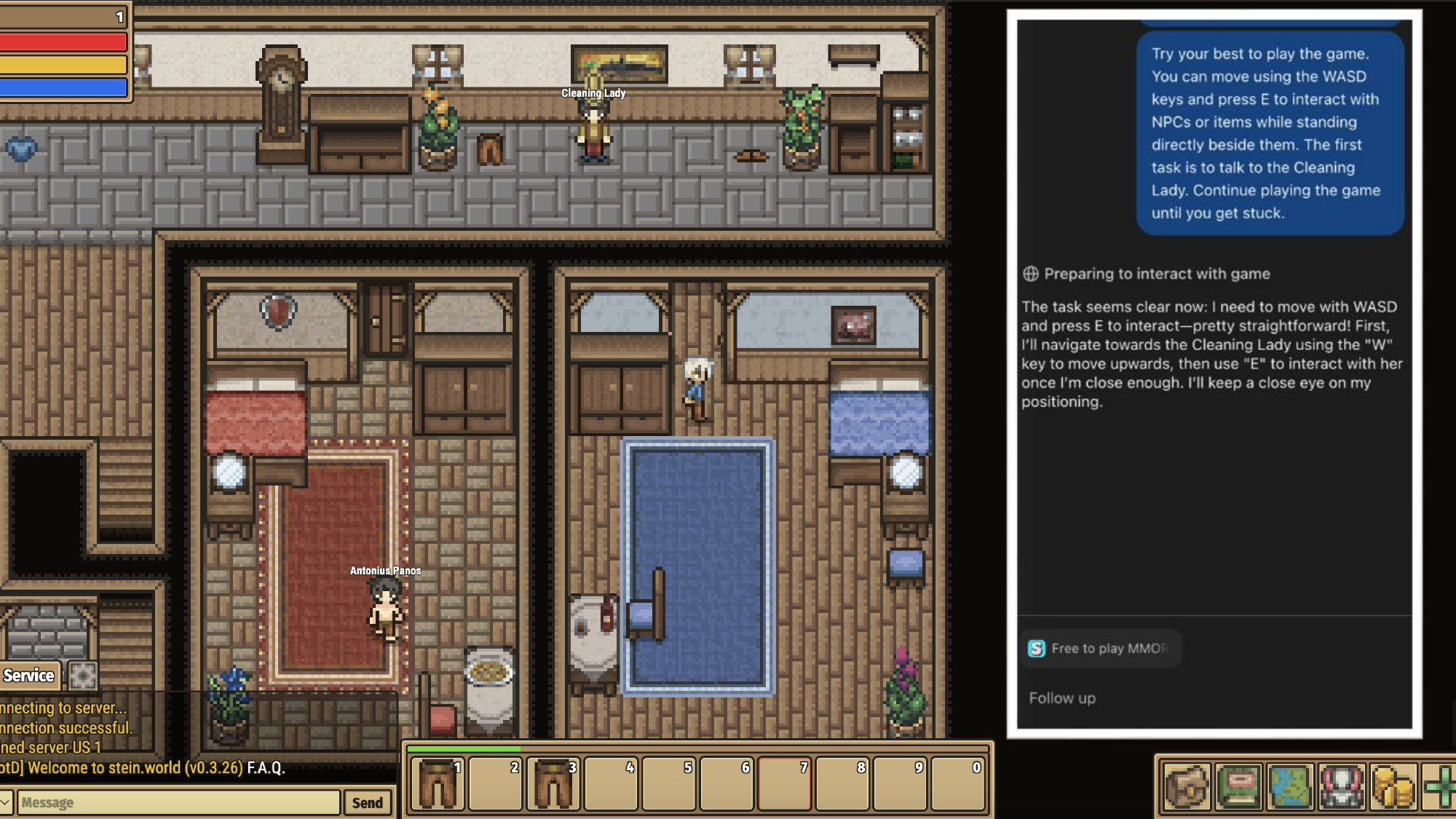}
    \caption{Layout of the Stein.world starting area and the location of the first task. The player begins in the spawn room and must navigate outside to speak with the NPC Cleaning Lady to advance the storyline. Atlas initially failed to exit the room despite extended attempts, but after receiving detailed operational instructions, it successfully completed the task.
    }
    \label{fig:stein}
\end{figure*}

\textbf{Strategy Analysis}: Atlas employed a systematic approach:
\begin{enumerate}
\item Complete board analysis and constraint identification
\item Mathematical modeling of possible number placements
\item Sequential input of numbers without hesitation or correction
\item Direct cursor selection of cells followed by number key presses
\end{enumerate}

The model's performance indicates sophisticated pattern recognition and logical deduction capabilities when freed from real-time motor coordination demands.

\paragraph{2048: Control Exploration with Minimal Planning}
In 2048, Atlas demonstrated a distinct \textbf{exploration-then-execution pattern} but achieved minimal strategic progress, typically reaching only the 64-tile before stalling.

\textbf{Observed Interaction Sequence:}
\noindent 

1. \textit{Initial exploration phase:} \\
\indent (a) Mouse clicking on directional arrows;\\
\indent (b) Testing keyboard arrow keys; \\
\indent (c) Discovering WASD controls. \\
During this phase, Atlas moved deliberately and at \textbf{a noticeably slower pace,} suggesting an emphasis on learning how to operate the interface.

\noindent 2. \textit{Execution phase:} \\
\indent (a) Fixed loop: 10 iterations of [Up, Right, Down, Left];\\
\indent (b) State assessment pause;\\
\indent (c) Random directional moves until stuck. \\
Once control understanding was established, execution proceeded much faster—Atlas cycled through the fixed movement sequence with \textbf{high speed} across ten loops.

\noindent 3. \textit{Adaptation attempt:} \\
\indent (a)\textbf{ "New Game"} button click when progress stalled;\\
\indent (b) Repeat exploration with similar results.

Despite successful interface discovery, Atlas exhibited \textbf{no evidence of understanding the game’s core strategy} of corner consolidation and tile merging. The fixed movement patterns, rapid execution, and random subsequent moves suggest an absence of strategic planning or state-value assessment.

\paragraph{Stein.world: Textual Understanding and Task Execution in RPG games.}
After receiving the initial prompt, “Try your best to explore and play the game. Stop when you get stuck,” Atlas spent several minutes exploring how to operate the game, including moving the player and interacting with non-player characters (NPCs) or items. Initially, Atlas attempted to move the character by left- or right-clicking. It later discovered that movement was controlled via the WASD keys.

The first in-game task required the player to speak with an NPC named Cleaning Lady to advance the storyline. This character is located outside the room where the player initially spawns, as illustrated in Figure~\ref{fig:stein}. The player must therefore navigate out of the starting room and interact with the NPC to continue. However, \textbf{Atlas spent more than twenty minutes attempting to exit the room and ultimately failed to do so.} Notably, even after discovering that movement could be performed using the keyboard, Atlas continued to intermittently use mouse clicks for navigation and interaction. Throughout most of the gameplay session, Atlas spent significant time deliberating on what action to take next—such as deciding whether to press WASD—before finally returning control to the user after being unable to exit the room.

To further investigate Atlas’s ability to play a role-playing game (RPG), we provided a more detailed prompt:

\textit{“Try your best to play the game. You can move using the WASD keys and press E to interact with NPCs or items while standing directly beside them. The first task is to talk to the Cleaning Lady. Continue playing the game until you get stuck.”}

As expected, with more explicit instructions, Atlas adapted more quickly to the game mechanics. Within eight minutes, it successfully navigated the player outside the room and picked up an item of clothing (pants) from the floor, although \textbf{it misidentified the item as a stool}. After approximately fifteen minutes, Atlas completed the first task of speaking with the Cleaning Lady and was subsequently assigned the next quest: to find a used shirt in the west room. Despite another fifteen minutes of play, Atlas made limited progress in exploration, as it spent considerable time deliberating on each next action rather than efficiently executing movement commands. Eventually, after failing to locate the used shirt, Atlas returned control to the user.


\subsection{Summary of Behavioral Patterns}

The consistent patterns across game types reveal fundamental characteristics of Atlas's web interaction capabilities:

\begin{itemize}
\item \textbf{Motor Control Gap}: Significant limitations in timing precision and continuous control.
\item \textbf{Analytical Strength}: Superior performance in logical reasoning and systematic problem-solving.
\item \textbf{Instruction Dependence}: Heavy reliance on explicit operational guidance, with limited capacity for inferring objectives from contextual narrative.
\item \textbf{Adaptive Intent}: Awareness of limitations manifested through control frequency adjustments and setting modifications.
\item \textbf{Strategic Deficiency}: Interface exploration without developing sophisticated game strategies.
\end{itemize}

These observations suggest that while ChatGPT Atlas possesses advanced analytical capabilities for structured tasks, it faces substantial challenges in dynamic environments requiring precise motor coordination, real-time adaptation, and contextual understanding. The model's performance demonstrates excellence in rule-based analytical tasks while struggling with open-ended environments that require narrative comprehension, implicit objective inference, and sustained goal-directed behavior.



\paragraph{Operational Challenges}
Several games revealed fundamental operational limitations of Atlas. In 2048, Atlas did not demonstrate the expected ability to develop a strategy for determining the optimal next step. Instead, it spent time exploring how to operate the game—initially attempting to click tiles to make moves. After discovering that movements could be performed using the W, A, S, and D keys, Atlas executed approximately ten random loops of “swirling” sequences, each consisting of four directional moves, totaling about forty movements. Surprisingly, after these iterations, Atlas did not attempt to analyze the board state or formulate a new strategy for subsequent moves. Rather, it repeated another set of ten random swirling loops. The overall progress was slow, with the best performance being a 512-tile achieved across ten trials before getting stuck or returning control. These observations highlight Atlas’s difficulties with goal comprehension and sequential decision-making in the context of playing web-based games.


\subsection{Limitations and Discussion}

Our observations revealed two main patterns: strong performance in analytical tasks like Sudoku contrasted with significant challenges in real-time control and precise operation. The consistent failures in reflex-based games suggest limitations in motor control and timing precision, while the operational difficulties in games like 2048 and Stein.world indicate challenges with basic interface manipulation.

It is worth noting that these games were not designed as comprehensive evaluation benchmarks, but rather as scenarios for observing web interaction behaviors. The small sample size and limited trial counts mean these observations should be considered preliminary. However, the consistent patterns across different game types provide useful initial insights into Atlas's current capabilities and limitations for interactive web tasks.

\section{Conclusion}

Our empirical evaluation of ChatGPT Atlas's web interaction capabilities across five game types reveals a system with notable analytical strengths but significant limitations in execution, adaptation, and contextual understanding. The model demonstrates robust performance in static, turn-based environments requiring logical reasoning and systematic analysis—excelling in puzzle games like Sudoku and showing competent strategic understanding in 2048. However, Atlas exhibits substantial limitations across multiple dimensions: in dynamic, real-time environments requiring low-latency motor control, in strategic planning for tile-based games like 2048, and crucially, in narrative comprehension and objective pursuit in RPG environments. The performance gap between analytical tasks and real-time execution challenges suggests that while Atlas represents a significant advancement in web-based AI assistance, it has not yet achieved generalized proficiency across the full spectrum of interactive web tasks. These findings indicate that current browser control capabilities, while effective for information retrieval and structured task completion, remain insufficient for applications requiring precise motor coordination, strategic planning in dynamic systems, and nuanced comprehension of narrative context. 

\section{Future Work}

This study establishes an initial framework for evaluating web interaction agents through game performance metrics. We plan to expand our evaluation to include a broader range of web applications beyond games, including dynamic web forms, interactive data visualizations, and complex web-based tools. Next, our analysis will extend to comparative evaluation across multiple web interaction agents, including specialized web automation tools and emerging multimodal systems, to better contextualize Atlas's capabilities within the broader landscape. We also intend to develop more sophisticated testing protocols that can disentangle the various components of web interaction—visual analysis, decision-making, and motor execution—to more precisely identify failure points. Finally, we will investigate the potential of targeted training approaches and architectural improvements to enhance real-time performance.

\bibliography{custom}




    
    
    
    

\end{document}